\pdfoutput=1

\documentclass[11pt]{article}

\usepackage{acl}

\usepackage{times}
\usepackage{latexsym}
\usepackage[T1]{fontenc}
\usepackage[utf8]{inputenc}
\usepackage{microtype}
\usepackage{inconsolata}
\usepackage{pdfpages}
\usepackage{multirow}
\usepackage{amsfonts}
\usepackage{amssymb,amsmath,amsthm,enumitem}
\usepackage{pifont}
\usepackage{enumitem}
\usepackage{soul}
\usepackage{color}
\usepackage{caption}
\usepackage{subcaption}

\usepackage{float}
\restylefloat{table}

\setlist[itemize]{noitemsep, topsep=0pt, partopsep=0pt, leftmargin=*}

\title{MemeCLIP: Leveraging CLIP Representations for Multimodal Meme Classification}

  \author{Siddhant Bikram Shah\textsuperscript{1} \quad Shuvam Shiwakoti\textsuperscript{2} \quad Maheep Chaudhary\textsuperscript{3} \quad Haohan Wang\textsuperscript{4} \\
         \textsuperscript{1}Northeastern University, USA\\
         \textsuperscript{2}Delhi Technological University, India \\ 
         \textsuperscript{3}Nanyang Technological University, Singapore \\
         \textsuperscript{4}University of Illinois Urbana-Champaign, USA\\
}

\begin{document}
{\makeatletter\acl@finalcopytrue  
\maketitle}

\begin{abstract}
The complexity of text-embedded images presents a formidable challenge in machine learning given the need for multimodal understanding of multiple aspects of expression conveyed by them. While previous research in multimodal analysis has primarily focused on singular aspects such as hate speech and its subclasses, this study expands this focus to encompass multiple aspects of linguistics: hate, targets of hate, stance, and humor. We introduce a novel dataset PrideMM comprising 5,063 text-embedded images associated with the LGBTQ+ Pride movement, thereby addressing a serious gap in existing resources. We conduct extensive experimentation on PrideMM by using unimodal and multimodal baseline methods to establish benchmarks for each task. Additionally, we propose a novel framework MemeCLIP for efficient downstream learning while preserving the knowledge of the pre-trained CLIP model. The results of our experiments show that MemeCLIP achieves superior performance compared to previously proposed frameworks on two real-world datasets. We further compare the performance of MemeCLIP and zero-shot GPT-4 on the hate classification task. Finally, we discuss the shortcomings of our model by qualitatively analyzing misclassified samples. Our code and dataset are publicly available at: \href{https://github.com/SiddhantBikram/MemeCLIP}{https://github.com/SiddhantBikram/MemeCLIP}.
\end{abstract}

\begin{figure*}[!h]
	\begin{subfigure}{0.173\textwidth}
		\includegraphics[width=30mm, height= 35 mm]{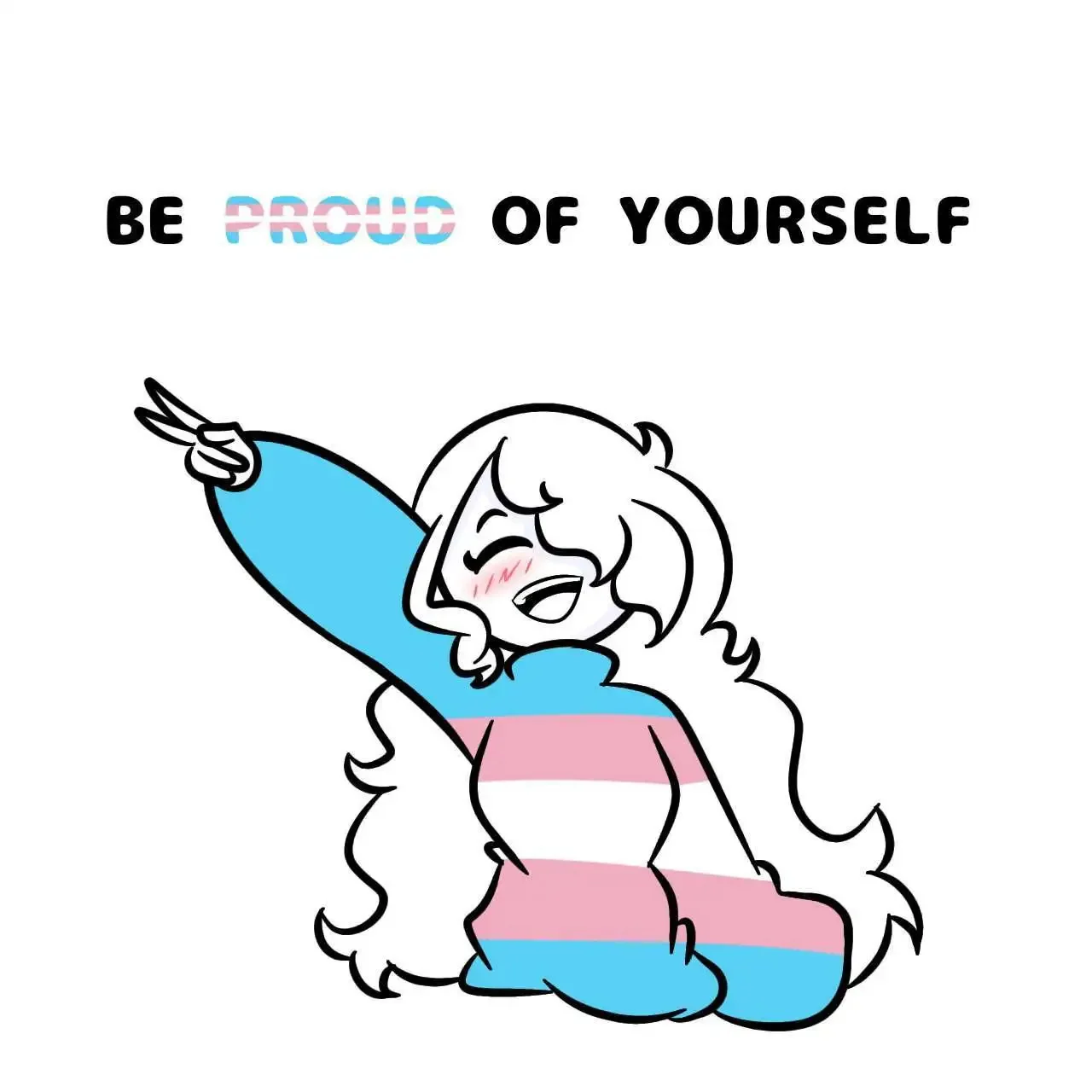}
		\caption{\{0, NA, 1, 0\}}\label{fig:a}		
	\end{subfigure}
	\hspace{0.6em}
        \begin{subfigure}{0.173\textwidth}
		\includegraphics[width=30mm, height= 35 mm]{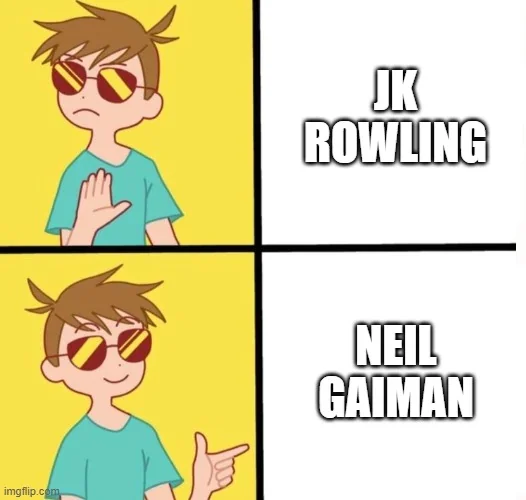}
		\caption{\{1, 1, 1, 1\}}\label{fig:b}		
	\end{subfigure}
	\hspace{0.6em}
	\begin{subfigure}{0.173\textwidth}
		 \includegraphics[width=30mm, height= 35 mm]{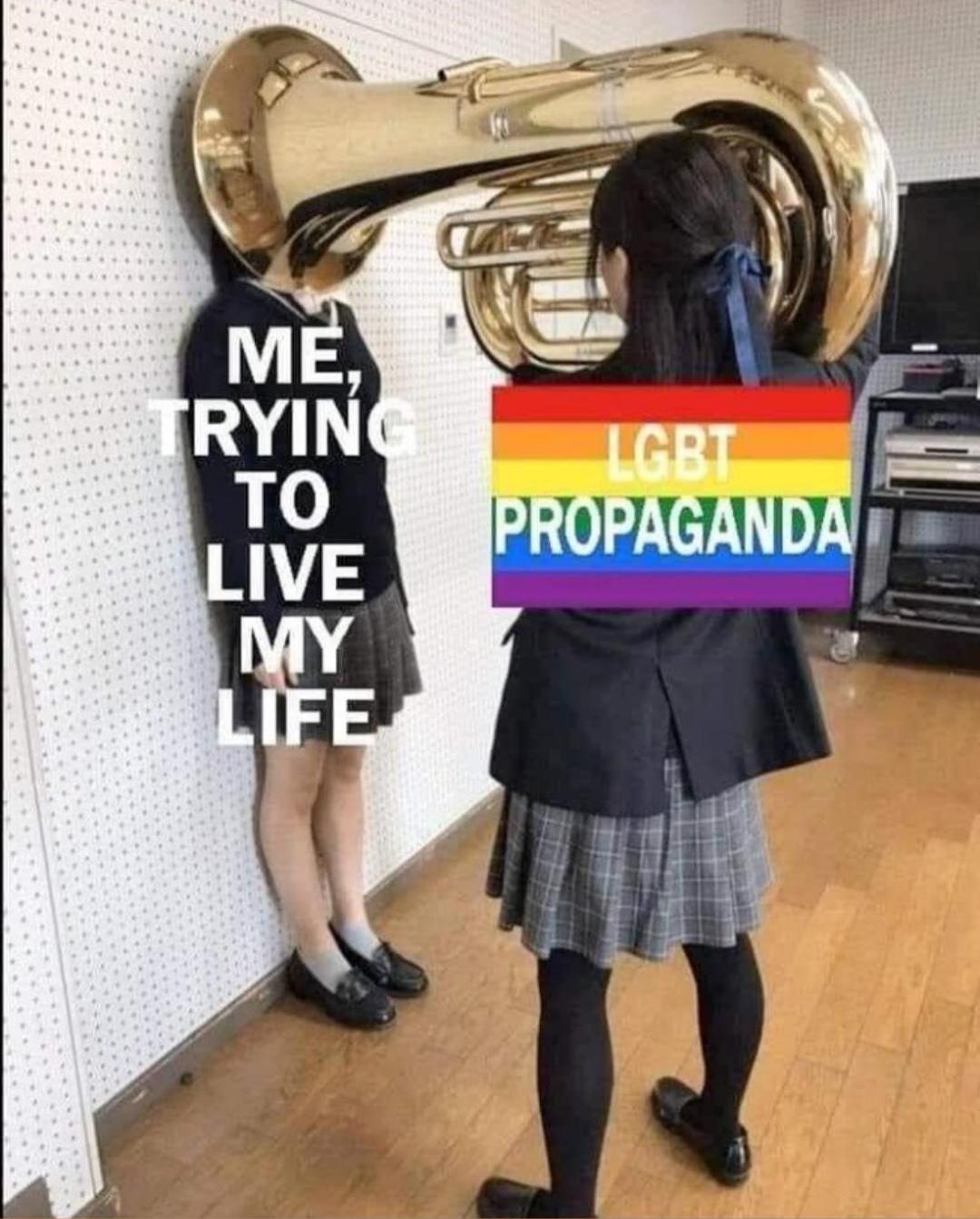}
		\caption{\{1, 0, 2, 1\}}\label{fig:c}
	\end{subfigure}
        \hspace{0.6em}
 	\begin{subfigure}{0.173\textwidth}
		\includegraphics[width=30mm, height= 35 mm]{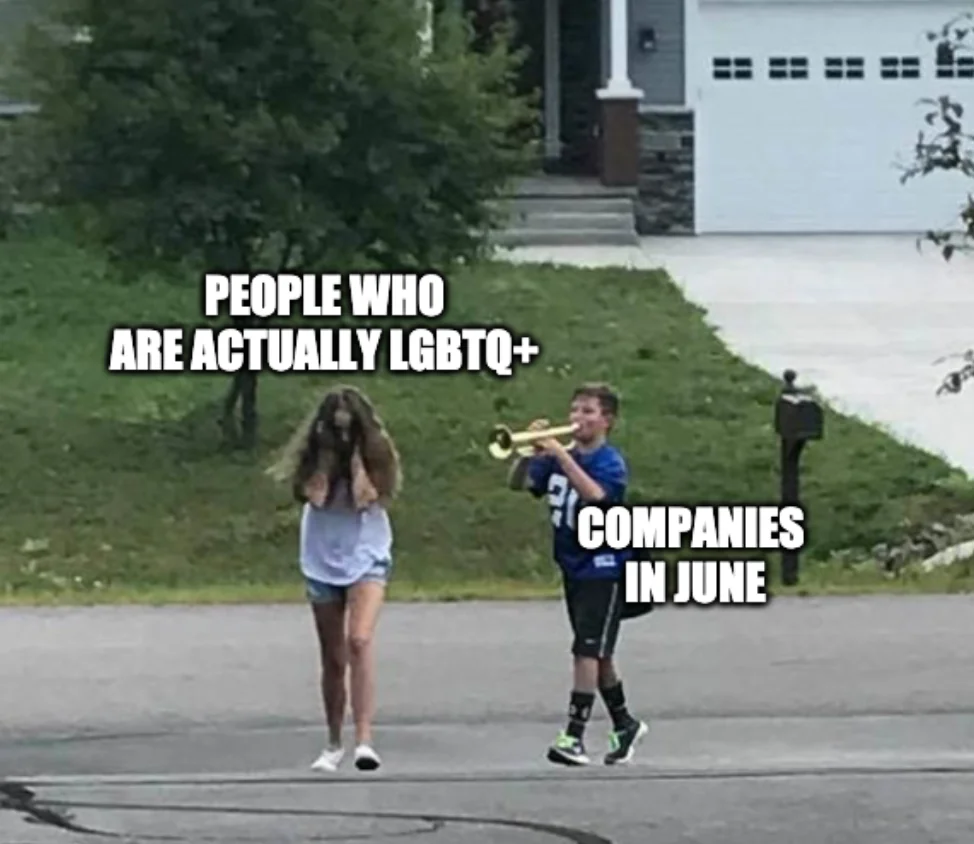}
		\caption{\{1, 3, 0, 1\}}\label{fig:d}		
	\end{subfigure}
        \hspace{0.6em}
  	\begin{subfigure}{0.173\textwidth}
		\includegraphics[width=30mm, height= 35 mm]{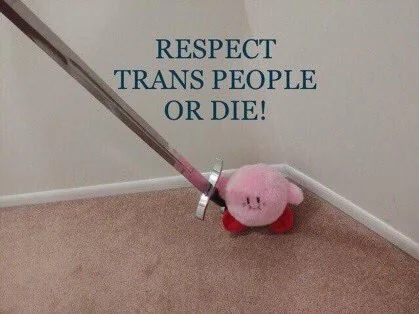}
		\caption{\{1, 0, 1, 1\}}\label{fig:e}		
	\end{subfigure}
	\caption{Samples of text-embedded images from the PrideMM dataset annotated across four aspect labels. The labels are in the form of \{Hate, Target, Stance, Humor\}. For Hate, \{0, 1\} correspond to \textit{No Hate} and \textit{Hate} respectively. For Target, \{0, 1, 2, 3\} correspond to hate targeted towards \textit{Undirected}, \textit{Individual}, \textit{Community}, and \textit{Organization} respectively. For Stance, \{0, 1, 2\} correspond to \textit{Neutral}, \textit{Support}, and \textit{Oppose} respectively. For Humor, \{0, 1\} correspond to \textit{No Humor} and \textit{Humor} respectively.}
 \label{fig:examples_1}
\end{figure*}

\section{Introduction}

In recent years, the pervasive integration of social media platforms into everyday life has resulted in an exponential increase in the generation and dissemination of multimedia content. At the heart of this digital ecosystem lies the meme: a text-embedded image imbued with humor, wit, and often, a subversive edge, which offers a medium through which individuals can express opinions, share experiences, and engage in online activism \cite{moreno2021memes, baker2020putin}. With their ability to distill complex ideas into digestible units of communication, memes have emerged as a powerful medium for expressing both support and opposition toward socio-political events \cite{imperato2023all}. 

However, with opinions being expressed freely, hate speech becomes prevalent, often directed towards individuals, organizations, and even marginalized communities \cite{thapa2023nehate}, targeting them with vitriol and prejudice \cite{lingiardi2020mapping,imperato2023all}. Particularly, the LGBTQ+ movement stands as a prominent subject of online discourse, where memes serve as vehicles of both solidarity and resistance, reflecting the multifaceted dynamics of attitudes and perceptions within the community and beyond \cite{gal2016gets}. In this context, the distinction between humor and harm becomes blurred, as memes straddle the line between satire and offense, challenging researchers and platforms alike to navigate the complexities of online content moderation \cite{langvardt2017regulating}. Previous attempts that endeavored to suppress such content have resulted in the discriminative suppression of all LGBTQ+ content \cite{griffin2022sanitised, griffin2024heteronormative}, which can harm the awareness and acceptance of this community. Thus, understanding the nuances of hate speech, opinions, and intended humor within memes becomes paramount for fostering an inclusive digital environment and combating online discrimination. 

To address these challenges, we introduce PrideMM: a novel dataset comprising 5,063 text-embedded images related to the LGBTQ+ movement annotated with a multi-aspect schema encompassing four tasks:

\begin{itemize}
    \item \textbf{Task A}: Detection of Hate Speech 
    \item \textbf{Task B}: Classifying the Targets of Hate Speech
    \item \textbf{Task C}: Classification of Topical Stance
    \item \textbf{Task D}: Detection of Intended Humor
\end{itemize}

The analysis of text-embedded images is particularly challenging given the need for contextual understanding and the prevalence of ambiguity and subjectivity in them \cite{sherratt2022towards}. Accordingly, the multi-aspect nature of our dataset provides a more holistic view of the diverse themes usually expressed through memes. Through PrideMM, we aim to cultivate a more profound understanding of interactions on social media through memes and facilitate the development of multimodal content moderation methods to make the internet a safer space. We implement a range of baseline and state-of-the-art hate speech detection models to establish benchmarks for each task of PrideMM. Sample images from PrideMM are illustrated in Figure \ref{fig:examples_1} alongside their annotation labels.

We further propose MemeCLIP, a novel framework that leverages the knowledge of the Contrastive Language-Image Pre-Training (CLIP) model \cite{radford2021learning} by using multiple lightweight modules for multimodal and multi-aspect meme classification. We employ linear layers to effectively disentangle image and text representations in CLIP's multimodal embedding space. We utilize Feature Adapters to preserve the prior knowledge of CLIP and adapt its embedding spaces to the meme classification task while avoiding overfitting on smaller datasets. We further implement a cosine classifier alongside Semantic-Aware initialization \cite{shi2023parameter} to make it more robust to the class imbalances that may exist in datasets such as PrideMM and HarMeme \cite{pramanick2021detecting} that are representative of real-world data distributions. Distinct from previous multimodal meme classification frameworks, MemeCLIP is trained end-to-end in a single step and does not rely on extraneous models to create augmented data. Our main contributions can be summarized as follows:

\begin{itemize}
    \item We release PrideMM, a dataset containing 5,063 text-embedded images related to the LGBTQ+ movement.
    \item We benchmark PrideMM by using various unimodal and multimodal methods including existing multimodal frameworks proposed for meme classification.
    \item We introduce MemeCLIP, a novel framework that utilizes lightweight modules on top of a frozen CLIP model to classify memes.
\end{itemize}

\renewcommand{\arraystretch}{1.2}

\begin{table*}[h]
\centering
\scriptsize
\caption{Summary of datasets used in the literature.}
\begin{tabular}{c c c c c c c}
\hline \hline
\small \textbf{Work} & \small \textbf{Data Source}& \small \textbf{Multimodal} & \small \textbf{Sub-Classes} & \small \textbf{Multi-aspect} & \small \textbf{Size} & \small \textbf{Context}\\
\hline \hline
\citet{qu2022disinfomeme} & Reddit & \ding{51} & \ding{55} & \ding{55} & 1,170 & COVID-19, BLM, Veganism \\
\citet{tanaka2022learning} & Meme Websites & \ding{51} & \ding{55} & \ding{55} & 7,500 & General Discourse \\
\citet{kiela2020hateful} & Self-Generated & \ding{51} & \ding{55} & \ding{55} & 10,000 & General Discourse \\
\citet{suryawanshi2020multimodal} & FB, Twitter, Instagram & \ding{51} & \ding{55} & \ding{55} & 743 & U.S. Election \\
\citet{pramanick2021detecting} & Google Images & \ding{51} & \ding{51} & \ding{55} & 3,544 & COVID-19 \\
\citet{pramanick2021momenta} & Google Images & \ding{51} & \ding{51} & \ding{55} & 3,522 & U.S. Politics \\
\citet{bhandari2023crisishatemm} & Twitter, FB, Reddit & \ding{51} & \ding{51} & \ding{55} & 4,723 & Russia-Ukraine War \\

\citet{dacon2022detecting} & Reddit & \ding{55} & \ding{51} & \ding{51} & 9930 & LGBTQ+ Movement\\
\citet{gautam2020metooma} & Twitter & \ding{55} & \ding{55} & \ding{51} & 9937 & \#MeToo Movement \\
\citet{ousidhoum2019multilingual} & Twitter & \ding{55} & \ding{51} & \ding{51} & 13,000 & General discourse \\

\hline
\textbf{{PrideMM} (Ours)} & {FB, Twitter, Reddit} & \ding{51} & \ding{51} & \ding{51} & {5,063} & LGBTQ+ Movement\\
\hline \hline
\end{tabular}

\label{table:related}
\end{table*}

\section{Related Work}

\subsection{Multimodal Datasets}
Multimodal image-text analysis has seen significant strides in recent years owing to the widespread popularity and availability of image-text pairs across social media. With the increasing need for hate speech and offensive content detection, multimodal datasets for hate speech detection have seen a particular surge. One of the first datasets in this domain was the Hateful Meme Challenge (HMC) dataset \cite{kiela2020hateful}, containing synthetic memes designed to convey contrastive implications from the image and text modalities that target religion, race, disability, and sex. Similarly, the Harm-C \cite{pramanick2021detecting} and Harm-P \cite{pramanick2021momenta} datasets comprise memes related to the COVID-19 pandemic and US politics respectively that were annotated across three degrees of harmfulness and four subclasses of hate speech targets. \citet{bhandari2023crisishatemm} annotated samples for hate speech detection and target classification similarly, collecting text-embedded images related to the Russia-Ukraine conflict from Twitter, Facebook (FB), and Reddit. Tangentially, \citet{suryawanshi2020multimodal} employed extensive annotation guidelines to create the MultiOFF dataset for offensive content detection, consisting of memes collected from Reddit, Facebook, Twitter, and Instagram. In an effort to discern humor often expressed in memes, \citet{tanaka2022learning} created a humor detection dataset by proposing a pipeline to extract memes devoid of interpersonal influence on the perception of humor. To identify disinformative memes, \citet{qu2022disinfomeme} introduced the DisinfoMeme dataset that contains memes related to COVID-19, the Black Lives Matter (BLM) movement, and Veganism.

\subsection{Multi-aspect Datasets}
Online discourse on socio-political events is often imbued with a series of human emotions, leading researchers to study the numerous aspects of linguistics expressed in them. \citet{dacon2022detecting}
used comments collected from RedditBias \cite{barikeri2021redditbias} related to LGBTQ+ individuals and annotated each comment for the presence of Toxicity, Severe Toxicity, Obscene, Threat, Insults, and Identity Attacks. Similarly, \citet{gautam2020metooma} curated a dataset of tweets related to the \#MeToo movement in social media by annotating the tweets across five different aspects. Taking multi-aspect datasets one step further, \citet{ousidhoum2019multilingual}
compiled an extensive multi-aspect Twitter dataset with English, French, and Arabic samples, with each annotated for different aspects including hate and offensiveness. Table \ref{table:related} provides a detailed comparison of the datasets cited in this section.
 
Multi-aspect data helps better encompass the spectrum of human emotions that may be associated with social media interactions. Most multimodal datasets, while only focusing on a single aspect and its sub-classes, fail to encompass the complex dynamics of emotions expressed by the masses. Our work aims to address this gap by presenting a multimodal and multi-aspect dataset comprising three different aspects- hate, topical stance, and humor, and one subclass within hate: targets of hateful speech, to enable more nuanced studies of multimodal meme data through computational methods.

\subsection{Multimodal Frameworks}
Recent developments in large vision-language models have incited a wave of research in methods to tackle hate speech in text-embedded images. MOMENTA \cite{pramanick2021momenta} was one of the first frameworks proposed to incorporate CLIP's vision and language encoders for multimodal hate speech classification. It extracts regions of interest from image data and named entities from text data to combine them with CLIP representations by using cross-modal attention fusion. Similarly, HateCLIPper \cite{kumar2022hate} was proposed to better model cross-modal interactions between CLIP representations. Textual inversion \cite{gal2022image} has been used to integrate visual cues in the text representation space in frameworks such as ISSUES \cite{burbi2023mapping}. Recent works make use of image caption models to extract text captions from images and learn a single language processing model \cite{cao2023prompting, cao2023pro}. However, rather than relying on augmented data from extraneous models, our proposed framework MemeCLIP leverages the knowledge learned by CLIP's encoders during its pre-training step to process the rich multimodal information inside each image. Additionally, we use Feature Adapters alongside residual connections to prevent overfitting as annotated datasets for multimodal meme classification generally lack a high number of samples. We further utilize a cosine classifier to make MemeCLIP more robust to imbalanced data classes, which is prevalent in multi-label tasks in this domain.        
\section{Dataset}
In this section, we describe various aspects of our dataset including data collection, annotation guidelines, and dataset statistics. Our dataset comprises 5,063 text-embedded images that encompass memes, posters, and infographics relevant to the LGBTQ+ movement. We only include images from 2020-2024 as this period saw an upsurge of social media content in this domain \cite{oz2023under}. This also allows our dataset to represent contemporary social media interactions through memes. Note that by the term LGBTQ+, we refer to all gender identities and sexual orientations inclusively.  

\subsection{Data Collection}
To maintain diversity in the dataset, we collected data from three popular social media platforms: Facebook, Twitter, and Reddit, through manual search and extraction. For Twitter, we used hashtags such as \textit{\#lgbt}, \textit{\#pride}, \textit{\#trans}, \textit{\#transrights}, \textit{\#nonbinary}, and \textit{\#genderidentity} to filter images related to LGBTQ+ discussions. For Facebook, we targeted groups that frequently discussed LGBTQ+ content. Similarly, for Reddit, we identified subreddits where discussion related to LGBTQ+ was more prominent. Further, to ensure the relevance and quality of the dataset, the data collection process was subject to filtering criteria. Detailed filtering criteria for our dataset can be found in Appendix \ref{apx:filter}. As different annotators may encounter and collect the same image, we sequentially employed two image deduplication tools: dupeGuru\footnote{https://github.com/arsenetar/dupeguru} and difPy\footnote{https://github.com/elisemercury/Duplicate-Image-Finder}, to search for duplicates and retain the highest quality image out of each batch of duplicates. We used the OCR application provided by Google Cloud Vision API\footnote{https://cloud.google.com/vision/docs} to extract textual data from the images. We removed non-alphanumeric elements such as special characters, hyperlinks, symbols, and non-English characters to reduce noisy text data and ensure data quality. Note that the text may occasionally contain unintentional noisy artifacts.

\subsection{Data Annotation}

We engaged five experienced annotators well-versed in NLP and computational linguistics to annotate data samples for PrideMM. The annotators had a prior understanding of the LGBTQ+ movement and meme archetypes on social media. We presented them with comprehensive annotation guidelines to ensure uniform and unbiased annotations, and asked them to annotate each image separately for all four tasks. A 3-phase annotation schema was used to ensure accurate and consistent annotations. First, a dry run was conducted to evaluate the understanding of the annotation guidelines among the annotators where every annotator was given an identical batch of 50 images for annotation. Second, a revision phase was conducted where every annotator was given another identical batch of 200 images and received a revised set of instructions based on the results of the first phase. Finally, in the consolidation phase, the annotators annotated a final batch of 50 images while discussing and revising the annotation guidelines until a consensus was reached. These steps were taken to minimize misannotations and noisy labels in the PrideMM dataset. The meticulously devised annotation guidelines were followed to ensure consistency in the annotations. Each image in our dataset was independently annotated for the three aspects and one sub-class, apart from the connection between 'Hate' and 'Hate Targets'.

\subsection{Annotation Guidelines}
\label{apx:guide}

In this section, we describe the annotation guidelines used to annotate the dataset. We devise separate guidelines for each of the four tasks.

\noindent \textbf{Hate Speech.} \quad This task aimed to identify instances of hate speech in the images. The primary focus was on identifying images that intentionally conveyed hateful sentiments. Annotators needed to distinguish between images expressing strong disagreement without resorting to offensive language and those containing genuine elements of hate speech. This differentiation aimed to guarantee accurate labeling, ensuring that images conveying genuinely hateful sentiment through visual content, language, or a combination of both were appropriately identified.

\noindent \textbf{Hate Targets.} \quad This task required annotators to identify the targets of hate in hateful images by classifying the images into one of the four classes: \textit{Undirected}, \textit{Individual}, \textit{Community}, and \textit{Organization}. Images were labeled as \textit{Undirected} when they targeted abstract topics, societal themes, or ambiguous targets like `you' that were not directed toward any specific individuals, entities, or groups. Hateful images targeting specific people including political leaders, celebrities, or activists like `Joe Biden' and `J.K. Rowling' were annotated as \textit{Individual}. Likewise, the label \textit{Community} was used for instances of images targeting broader social, ethnic, or cultural groups like `LGBT' or `trans'. Lastly, images targeting corporate entities, institutions, or similar organizations like `Chick-fil-A' and `government" were annotated as \textit{Organization}.

\noindent \textbf{Stance.} \quad This task involved annotating the images into either of three distinct categories: \textit{Support}, \textit{Oppose}, and \textit{Neutral}, determined by their stance within the context of the LGBTQ+ movement. The \textit{Support} label was given to images that expressed support towards the goals of the movement, agreed with efforts in fostering equal rights for LGBTQ+ individuals, and promoted awareness for the movement's goals. The \textit{Oppose} label was given to images that conveyed disagreement with the goals of the movement, denied the problems faced by individuals who identified as LGBTQ+, and dismissed the need for equal rights and acceptance. The \textit{Neutral} label was given to images that were contextually relevant to the movement but did not exhibit support or opposition towards the movement.

\noindent \textbf{Humor.} \quad In this task, annotators were asked to identify images showcasing humor, sarcasm, or satire related to the LGBTQ+ Pride movement. Annotators were instructed to discern the presence of humor in the images regardless of whether they presented a lighthearted or insensitive perspective on serious subjects. Note that annotators were asked to annotate images based on whether the creator of the image intended for it to be humorous, and not based on whether the annotator personally found it humorous. This task aimed to capture the nuanced use of text-embedded images for comedic or satirical purposes, thereby helping disentangle hate and humor in the images related to this movement. 

\subsection{Statistics and Inter-Annotator Agreement}

Table \ref{table:dataset} shows the distribution of images in PrideMM across all class labels. For the hate detection task, the dataset has a balanced distribution of binary labels. The target classification task exhibits a heavily imbalanced distribution. Given the context of this study, most hateful images convey undirected hate or are targeted toward communities, with a low frequency of hate against individuals and organizations. For the stance classification task, the number of images is well-balanced across three labels. On the other hand, as memes are often meant to be humorous, the majority of the images in the dataset are annotated to humor. We use topic modeling to analyze PrideMM's text content, and the results are presented in Appendix \ref{appendix:topicmodeling}.

We used the Fleiss' Kappa (${\kappa}$) \citep{falotico2015fleiss} as a statistical measure to assess the inter-annotator agreement across all four tasks. For Task A (Hate Speech detection), ${\kappa}$ was 0.66/0.74 in the dry run and final phase respectively, for Task B (Target detection), ${\kappa}$ was 0.68/0.81, for Task C (Stance detection), ${\kappa}$ was 0.62/0.75, and for Task D (Humor detection), ${\kappa}$ was 0.60/0.74. The increase in ${\kappa}$ from the dry run phase to the final phase across all tasks reflects the effectiveness of the 3-phase annotation schema.

\begin{table}[!htpb]
\setlength{\tabcolsep}{3pt}

\renewcommand{\arraystretch}{1.2}
\centering
\caption{Dataset Statistics for PrideMM. The data consists of 5,063 samples for Hate, Stance, and Humor classification tasks, and 2,482 samples for the Target classification task.}
\small
\begin{tabular}{ c  c  c  c}

\hline \hline
\textbf{Task} & \textbf{Label} & \textbf{\#Samples} & \textbf{\%}\\
\hline \hline
\multirow{2}{*}{{Hate}} & No Hate & 2,581 & 50.97\%\\
& Hate & 2,482 & 49.03\%\\
\hline
\multirow{4}{*}{{Target}} & Undirected & 771 & 31.07\%\\
& Individual & 249 & 10.03\%\\
& Community & 1,164 & 46.90\%\\
& Organization & 298 & 12.00\%\\
\hline
\multirow{3}{*}{{Stance}} & Neutral & 1,458 & 28.80\%\\
& Support & 1,909 & 37.70\%\\
& Oppose & 1,696 & 33.50\%\\
\hline
\multirow{2}{*}{{Humor}} & No Humor & 1,642 & 32.43\%\\
& Humor & 3,421 & 67.57\%\\
\hline \hline
\end{tabular}
\label{table:dataset}
\end{table}

\begin{figure*}[!htpb]
\centering
\includegraphics[width= \linewidth]{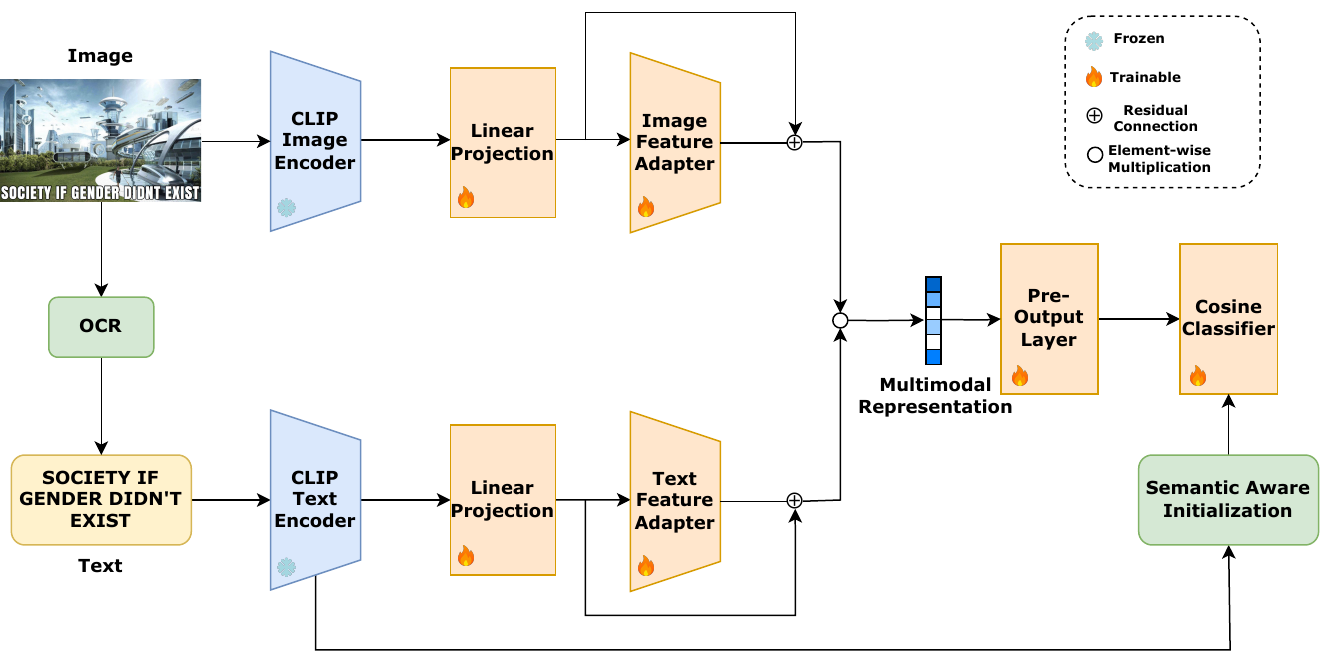}
\caption{An overview of our proposed framework, MemeCLIP. We use frozen CLIP image and text encoders to create representations for each image-text pair. These representations are passed through linear layers to disentangle the modalities in CLIP's shared embedding space. We implement Feature Adapters with residual connections for each modality to prevent overfitting. We use a cosine classifier to make MemeCLIP more robust to imbalanced data. We initialize classifier weights by using Semantic-Aware Initialization to further improve performance.}
\label{fig:MemeCLIP}
\end{figure*}

\section{Methodology}

In this section, we describe our proposed framework, MemeCLIP, for multi-aspect meme classification. We utilize the vision-language model CLIP \cite{radford2021learning} to create rich representations that effectively encapsulate the semantics of a meme. We add lightweight modules on top of CLIP to disentangle image and text representations, prevent overfitting, and make MemeCLIP more robust to imbalanced data. Figure \ref{fig:MemeCLIP} illustrates the overall architecture of MemeCLIP. Below, we describe each component of MemeCLIP in detail. 

\noindent \textbf{Zero-shot CLIP.} \quad The vision-language model CLIP exhibits stellar zero-shot performance and transfer learning capabilities \cite{radford2021learning, wang2023clipn}. CLIP is pre-trained on 400 million image-text pairs from the internet, enabling it to encode visual and textual data in a shared embedding space. The model consists of an Image Encoder $E_I$ and a Text Encoder $E_T$. We freeze the weights of both encoders to preserve the valuable knowledge captured by them during pre-training. The unimodal image and text representations $F_I, F_T \in \mathbb{R}^{768}$ effectively encapsulate the semantics of a meme and are defined as:

\vspace{-5pt}

\begin{equation} \label{eq1}
    F_I = E_I(I); \hspace{5pt} F_T = E_T(T)
\end{equation}

where I is the image and T is its text pair.

\noindent \textbf{Linear Projection Layers.} \quad While the contrastive pre-training objective of CLIP promotes similarity between corresponding text and image pairs, memes often involve contrastive visual and linguistic content to evoke a sense of irony. Similar to \cite{kumar2022hate}, we employ individual linear projection layers for each modality to effectively disentangle image and text representations in the shared embedding space. These projection layers result in the unimodal projections $F^{proj}_I, F^{proj}_T \in \mathbb{R}^{1024}$, mapping the representations to the dimensions of CLIP's last hidden state, $D_{CLIP} \in \mathbb{R}^{1024}$, which enables the use of Semantic-Aware Initialization.

\vspace{-10pt}

\begin{equation} \label{eq2}
    F^{proj}_I = L^{proj}_I(F_I); \hspace{5pt} F^{proj}_T = L^{proj}_T(F_T) 
\end{equation}

Here, $L^{proj}_I$ and $L^{proj}_T$ represent the image and text projection layers respectively. 

\noindent \textbf{Feature Adapters.} \quad Since CLIP is pre-trained on an extensive amount of data, it may exhibit symptoms of overfitting when applied to smaller datasets for downstream tasks. Inspired by \cite{gao2024clip}, we adopt lightweight Feature Adapters for both image and text modalities to learn the features of new data while retaining CLIP's prior knowledge. We further utilize residual connections to integrate prior image and text projections with the outputs of the Adapters, allowing our model to balance the knowledge of the fine-tuned adapter and the disentangled image and text projections. We use a residual ratio $\alpha$ to maintain harmony between these two modules. With the image and text Feature Adapters $A_I$ and $A_T$ respectively, the final unimodal representations $F_I, F_T \in \mathbb{R}^{1024}$ are obtained as follows: 

\vspace{-10pt}
\begin{equation} \label{eq3} 
    F_I = \alpha A_I(F^{proj}_I) + (1 - \alpha)F^{proj}_I 
\end{equation}

\vspace{-25pt}
\begin{equation} \label{eq4} 
    F_T = \alpha A_T(F^{proj}_T) + (1 - \alpha)F^{proj}_T 
\end{equation}


\noindent \textbf{Modality Fusion.} \quad Owing to the extensive unimodal feature modeling, MemeCLIP avoids the need for trainable fusion layers like Cross-Modal Attention Fusion in MOMENTA \cite{pramanick2021momenta} and Combiner in ISSUES \cite{burbi2023mapping}. We fuse the image and text representations by using an element-wise multiplication operation ($\circ$) to obtain a single multimodal representation $F_{MM} \in \mathbb{R}^{1024}$. We further alleviate the need for the two-stage training process employed in HateCLIPper \cite{kumar2022hate} and ISSUES. 

\vspace{-10pt}

\begin{equation} \label{eq5} 
    F_{MM} = F_I \circ F_T
\end{equation}

$F_{MM}$ is then passed through a linear pre-output layer before classification.


\noindent \textbf{Classification.} \quad For classification, we employ a cosine classifier \cite{liu2020deep} that is robust to biases in prediction under class imbalances. Following \cite{shi2023parameter}, we adopt Semantic-Aware Initialization (SAI) to initialize the weights of this classifier by exploiting the semantic knowledge held within the text encoder of CLIP. We encode class labels by using the prompt "A photo of \{LABEL\}" into $F_{class} \in \mathbb{R}^{n \times 1024}$ where n is the number of classes. We use $F_{class}$ to initialize the classifier weight $W_{class}$. During training, the predicted logit $Z$ for a class $x$ is calculated as follows:

\begin{equation} \label{eq6} 
    Z_x = \sigma \times  \frac{W_x \times F_{MM}}{||W_x||_2 \hspace{5pt} ||F_{MM}||_2}
\end{equation}

where $\sigma$ is a static scaling factor for the cosine classifier.

\begin{table*}[!h]
\renewcommand{\arraystretch}{1.6}
\setlength{\tabcolsep}{2pt}
\centering
\scriptsize

\caption{Classification performance of methods on the PrideMM dataset. The results are in the form of \textit{Mean ± Standard Deviation}. Performance is reported across three evaluation metrics: Accuracy, AUROC (Macro), and F1-Score (Macro). The best performance is highlighted in \textbf{bold}.}
\begin{tabular}{c | c c c | c c c | c c c | c c c}
\hline \hline
\multirow{2}{*}{\textbf{Method}} & \multicolumn{3}{c|}{\textbf{Hate}} & \multicolumn{3}{c|}{\textbf{Target}} & \multicolumn{3}{c|}{\textbf{Stance}} & \multicolumn{3}{c}{\textbf{Humor}}\\

& \textbf{Acc.} & \textbf{AUC.} & \textbf{F1} & \textbf{Acc.} & \textbf{AUC.} & \textbf{F1} & \textbf{Acc.} & \textbf{AUC.} & \textbf{F1} & \textbf{Acc.} & \textbf{AUC.} & \textbf{F1} \\
\hline \hline

BERT & 71.12\tiny{±0.67} & 75.33\tiny{±0.50} & 70.06\tiny{±0.22} & 54.25\tiny{±1.32} & 75.52\tiny{±0.77} & 54.03\tiny{±1.51} & 52.30\tiny{±0.91} & 67.10\tiny{±0.74} & 51.11\tiny{±1.34} & 71.04\tiny{±0.55} &72.25\tiny{±1.77} & 64.60\tiny{±0.98}\\

CLIP Text-Only & 68.64\tiny{±0.86} & 74.52\tiny{±0.95} & 68.62\tiny{±0.88} & 50.34\tiny{±0.62} & 72.67\tiny{±1.34} & 47.65\tiny{±1.18} & 50.43\tiny{±1.08} & 67.60\tiny{±0.55} & 49.26\tiny{±1.25} &	69.23\tiny{±1.54} &	70.52\tiny{±1.42} &	62.02\tiny{±1.69} \\

ViT-L/14 & 69.23\tiny{±2.22} & 77.05\tiny{±0.48} & 68.24\tiny{±3.01} & 58.36\tiny{±1.56} & 79.13\tiny{±1.48} & 50.24\tiny{±2.93} & 58.80\tiny{±1.26} & 73.20\tiny{±4.62}	& 56.14\tiny{±4.76} &	73.04\tiny{±2.20} &	77.71\tiny{±1.13} &	69.18\tiny{±2.35} \\ 
CLIP Img-Only & 70.01\tiny{±0.78} & 80.53\tiny{±0.42} & 72.66\tiny{±3.08} & 60.32\tiny{±1.77} & 80.89\tiny{±0.73} & 57.19\tiny{±3.25} & 61.01\tiny{±0.82} & 77.48\tiny{±0.66} & 57.87\tiny{±0.78} & 76.14\tiny{±0.19} & 82.1\tiny{±1.42} & 72.37\tiny{±1.13} \\

\hline

CLIP & 72.39\tiny{±1.20} & 80.47\tiny{±0.61} & 72.33\tiny{±1.26} & 61.14\tiny{±0.59} & 81.92\tiny{±0.44} & 58.46\tiny{±1.02} & 59.31\tiny{±0.82} & 76.92\tiny{±0.87} & 57.81\tiny{±1.14} & 76.66\tiny{±1.32} & 80.73\tiny{±0.20} & 73.23\tiny{±1.56} \\ 

CLIP-Adapter & 72.75\tiny{±1.09} & 80.91\tiny{±0.56} & 72.69\tiny{±1.02} & 61.59\tiny{±0.52} & \textbf{82.14}\tiny{±0.35} & 58.08\tiny{±0.91} & 59.55\tiny{±0.47} & 77.23\tiny{±0.73} & 57.93\tiny{±0.91} & 77.01\tiny{±1.01} & 80.97\tiny{±0.71} & 73.51\tiny{±0.97} \\ 

MOMENTA & 72.23\tiny{±0.58} & 78.55\tiny{±0.50} & 71.78\tiny{±0.35} & 57.28\tiny{±1.26} & 78.89\tiny{±1.23} & 52.79\tiny{±1.84} & 55.62\tiny{±1.90} & 73.64\tiny{±2.35} & 54.84\tiny{±2.28} &	74.16\tiny{±2.17} & 77.38\tiny{±1.63} & 71.34\tiny{±2.70} \\ 

HateCLIPper & 75.53\tiny{±0.58} & 83.12\tiny{±0.44} & 74.08\tiny{±0.37} & 62.49\tiny{±2.06} & 80.32\tiny{±1.42} & 56.77\tiny{±0.72} & \textbf{63.24}\tiny{±0.69} & 77.99\tiny{±1.25} & 57.15\tiny{±0.76} &	76.13\tiny{±0.19} & 83.50\tiny{±0.51} & 75.41\tiny{±0.28} \\

ISSUES & 74.68\tiny{±1.62} & 84.17\tiny{±0.45} & 73.64\tiny{±2.48} & 61.25\tiny{±2.00} & 78.73\tiny{±0.21} & 58.30\tiny{±0.17} & 59.39\tiny{±1.08} & 77.02\tiny{±1.93} & 57.27\tiny{±1.40} & 78.95\tiny{±0.88} & 84.78\tiny{±0.60} & 75.73\tiny{±2.17} \\

\hline
MemeCLIP & \textbf{76.06}\tiny{±0.23} & \textbf{84.52}\tiny{±0.31} & \textbf{75.09}\tiny{±0.20} & \textbf{66.12}\tiny{±0.47} & 81.66\tiny{±0.25} & \textbf{58.65}\tiny{±0.97} & 62.00\tiny{±0.12} & \textbf{80.11}\tiny{±0.15} & \textbf{57.98}\tiny{±1.91} & \textbf{80.27}\tiny{±0.52} & \textbf{85.59}\tiny{±0.23} & \textbf{77.21}\tiny{±0.79} \\
\hline \hline

\end{tabular}
\label{table:Results_Table}

\end{table*}

\section{Experimental Results}

Table \ref{table:Results_Table} and Table \ref{table:Results_Table_harmc} show our experimental results for the PrideMM and HarMeme datasets respectively. We pre-define train/validation/test splits in the ratio 85/5/10 respectively for PrideMM, and use the pre-defined split for the HarMeme dataset. We conduct experiments on unimodal and multimodal baseline methods, and previous frameworks proposed for multimodal meme classification. We conduct each experiment on three random seeds and report the Mean and Standard Deviation (±) values for Accuracy, AUC (Macro), and F1-Score (Macro). We use ViT-L/14 as the image encoder for all CLIP-based methods except for MOMENTA, which uses ViT-B/32 as the backbone for its CLIP model by default. For CLIP, we use concatenation to fuse the unimodal feature representations. Further implementation details are outlined in \ref{appendix:implement}. 

\subsection{PrideMM Dataset}

\textbf{Unimodal Methods.} \quad For the unimodal methods, we used ViT-L/14 \cite{dosovitskiy2020image} and CLIP's image encoder as image-based methods, and BERT \cite{devlin2018bert} and CLIP's text encoder as text-based methods. The image-based methods generally performed better than their text-based counterparts across all tasks, substantiating that Transformer-based visual models create meaningful representations that also capture the semantic meaning conveyed by the text embedded in the pixel space \cite{burbi2023mapping}. The text-based methods showed poor performance on the multi-label target and stance detection tasks. The image encoder of CLIP shows superior results compared to the standard pre-trained Visual Transformer while having the same architecture, demonstrating the effectiveness of contrastive pre-training. The unimodal methods generally perform worse than any multimodal method across all tasks, underscoring the need for multimodal processing in meme analysis.

\noindent \textbf{Multimodal Methods.} \quad  We tested the performance of the multimodal methods CLIP \cite{radford2021learning}, CLIP-Adapter \cite{gao2024clip}, MOMENTA \cite{pramanick2021momenta}, HateCLIPper \cite{kumar2022hate}, ISSUES \cite{burbi2023mapping}, and our framework, MemeCLIP. MemeCLIP outperforms the baseline CLIP model and previously proposed multimodal methods across all metrics in the hate and humor classification tasks. Our model performs particularly well in the four-class target classification task, which has less than half the number of samples as the other tasks and harbors a heavy class imbalance, demonstrating the model's robustness to overfitting on majority classes in imbalanced datasets. Smaller frameworks with a lower number of parameters such as the baseline CLIP, CLIP-Adapter, HateCLIPper, and MemeCLIP perform optimally in this task with CLIP-Adapter showing the highest AUC, while the remaining methods show symptoms of overfitting. In the stance classification task, HateCLIPper surpasses MemeCLIP in accuracy, while the latter shows a higher AUC and F1-Score. While the CLIP-Adapter model is similar to MemeCLIP, it uses a single feature adapter as individual feature adapters for both modalities may carry redundant information from the unimodal encoders to the classifier; however, MemeCLIP outperforms this model in most tasks by using feature adapters for both image and text modalities, which may signify that separate layers for each modality are more effective at encoding memes that may contain visual and language content that convey different meanings individually and combined. 

\begin{table}[h]
\renewcommand{\arraystretch}{1.3}
\centering
\small

\begin{tabular}{c | c c c }
\hline \hline
\textbf{Method} & \textbf{Acc.} & \textbf{AUC.} & \textbf{F1} \\

\hline \hline

BERT & 71.05\tiny{±0.70} & 76.34\tiny{±0.48
} & 68.83\tiny{±0.49} \\

CLIP Text-Only & 73.79\tiny{±0.23} & 79.31\tiny{±0.81} & 71.60\tiny{±0.21} \\

ViT-L/14 & 77.27\tiny{±1.71} & 85.53\tiny{±0.41} & 75.55\tiny{±2.63} \\

CLIP Img-Only & 79.38\tiny{±0.25} & 88.54\tiny{±2.53} & 78.66\tiny{±0.11} \\

\hline

CLIP & 81.36\tiny{±0.81} & 87.27\tiny{±0.67} & 80.30\tiny{±0.95} \\

CLIP-Adapter & 82.21\tiny{±0.73} & 87.53\tiny{±0.61} & 80.89\tiny{±0.87} \\

MOMENTA & 82.44\tiny{±0.65} & 87.88\tiny{±0.37} & 81.49\tiny{±0.45} \\

HateCLIPper & 83.68\tiny{±0.62} & 90.83\tiny{±0.46} & 83.31\tiny{±0.41} \\

ISSUES & 81.31\tiny{±1.05}  & 91.98\tiny{±0.61} & 80.45\tiny{±0.87} \\

\hline
MemeCLIP & \textbf{84.72}\tiny{±0.45}  & \textbf{92.07}\tiny{±0.34} & \textbf{83.74}\tiny{±0.43}  \\

\hline \hline

\end{tabular}
\caption{Classification performance of methods on the HarMeme dataset. The results are in the form of \textit{Mean ± Standard Deviation}. Performance is reported across three evaluation metrics: Accuracy, AUROC (Macro), and F1-Score (Macro). The best performance is highlighted in \textbf{bold}.}
\label{table:Results_Table_harmc}

\end{table}

\subsection{HarMeme Dataset}

To test the generalizability of MemeCLIP across other meme datasets, we perform experiments on the HarMeme dataset \cite{pramanick2021detecting}, which consists of real-world hateful memes shared on social media in the context of COVID-19. Similar to previous studies, we find that the performance of the multimodal models surpasses the unimodal models' performance by a decent margin. MemeCLIP outperforms the other multimodal frameworks in this dataset, demonstrating its effectiveness over the state-of-the-art baselines. 

\subsection{Ablation Study}

We conduct a systematic ablation study to assess the contribution of each component of MemeCLIP towards its performance. The results of our ablation experiments are presented in Table \ref{table:ablation}. We start with the CLIP ViT-L/14 model and gradually integrate each external module of MemeCLIP. The rise in performance when the projection layers are applied signifies the importance of adapting CLIP's embedding spaces to our downstream task by disentangling the image and text representations. While the introduction of Feature Adapters initially leads to a temporary dip in F1-Score, it ultimately enables our model to produce more refined image and text representations due to the added learnable parameters. Replacing the linear classifier with a cosine classifier boosts performance by modulating weight updates with a static scaling factor. Finally, Semantic-aware initialization completes MemeCLIP by initializing classifier weights according to the semantic differences in class labels encoded by CLIP's text encoder, enhancing generalization further.

\begin{table}[!htpb]
\setlength{\tabcolsep}{3pt}
\renewcommand{\arraystretch}{1.3}
\centering
\caption{Ablation experiments performed on MemeCLIP using the hate detection task of the PrideMM dataset. The results are in the form of \textit{Mean ± Standard Deviation}. PL, FA, CC, and SAI denote Projection Layers, Feature Adapters, Cosine Classifier, and Semantic-Aware Initialization respectively. The last line represents the complete framework. The best results are highlighted in \textbf{bold}.}
\small
\begin{tabular}{ c  c  c c  c c c c}

\hline \hline
\textbf{CLIP} & \textbf{PL} & \textbf{FA} & \textbf{CC} & \textbf{SAI} & \textbf{Acc.} & \textbf{AUC.} & \textbf{F1} \\
\hline \hline

\ding{51} & & & & & 72.39\tiny{±0.61} & 80.47\tiny{±1.20} & 72.33\tiny{±1.26} \\
\ding{51} & \ding{51} &&&& 74.66\tiny{±0.03} & 81.68\tiny{±0.36} & 75.04\tiny{±0.45}  \\
\ding{51} & \ding{51} & \ding{51}  &&& 75.33\tiny{±0.17} & 83.44\tiny{±0.01}  & 74.77\tiny{±0.88}  \\
\ding{51} & \ding{51} & \ding{51}  & \ding{51} && 75.78\tiny{±0.32} & 84.35\tiny{±0.05}  & 74.92\tiny{±0.38}  \\
\hline
\ding{51}  & \ding{51} & \ding{51}  & \ding{51}  & \ding{51}  & \textbf{76.06}\tiny{±0.23} & \textbf{84.52}\tiny{±0.31} & \textbf{75.09}\tiny{±0.20} \\

\hline \hline
\end{tabular}
\label{table:ablation}
\end{table}

\subsection{Comparison with GPT-4}

Table \ref{table:gpt} compares the performance of MemeCLIP against zero-shot GPT-4 \cite{achiam2023gpt} through Microsoft Copilot\footnote{https://copilot.microsoft.com/} (accessed July 2024). We used the prompt \textit{"Is this image hateful or not? Consider if the image and its text are hateful towards individuals, communities, organizations, or an undirected target. Also, consider the context of the entities represented in the image. Reply with only a number, 0 for no, and 1 for yes."} to manually test GPT-4's performance on the hate classification task for 100 images each from the PrideMM and HarMeme dataset's test set. We qualitatively found that while GPT-4 showed stellar zero-shot performance for hate classification, it tends to make cautious predictions by classifying non-hateful images as hateful or unsafe to the detriment of performance. This behavior may be caused by the stringent safety measures applied to commercial LLMs by LLM providers \cite{korbak2023pretraining, bai2022constitutional}.

\begin{table}[h]
\setlength{\tabcolsep}{3pt}
\renewcommand{\arraystretch}{1.4}

\centering
\caption{Performance comparison between MemeCLIP and the GPT-4 model provided by Microsoft Copilot on hate classification for PrideMM and harm classification for HarMeme. The best results are highlighted in \textbf{bold}.} 

\small
\begin{tabular}{ c | c c c | c c c}

\hline \hline
\multirow{2}{*}{\textbf{Method}} & \multicolumn{3}{c|}{\textbf{PrideMM}} & \multicolumn{3}{c}{\textbf{HarMeme}}\\

&\textbf{Acc.} & \textbf{AUC.} & \textbf{F1} & \textbf{Acc.} & \textbf{AUC.} & \textbf{F1}\\
\hline \hline

GPT-4 & 70.00 & - & 69.38 & 78.00 & - & 74.41\\

MemeCLIP & \textbf{73.00} & \textbf{79.06} & \textbf{72.54} & \textbf{85.00} & \textbf{92.58} & \textbf{83.02}\\

\hline \hline
\end{tabular}
\label{table:gpt}
\end{table}

\subsection{Misclassification Analysis}

\begin{figure}[!h]
\centering

	\begin{subfigure}{0.16\textwidth}
            \captionsetup{justification=centering}
		\includegraphics[width=32mm, height= 32mm]{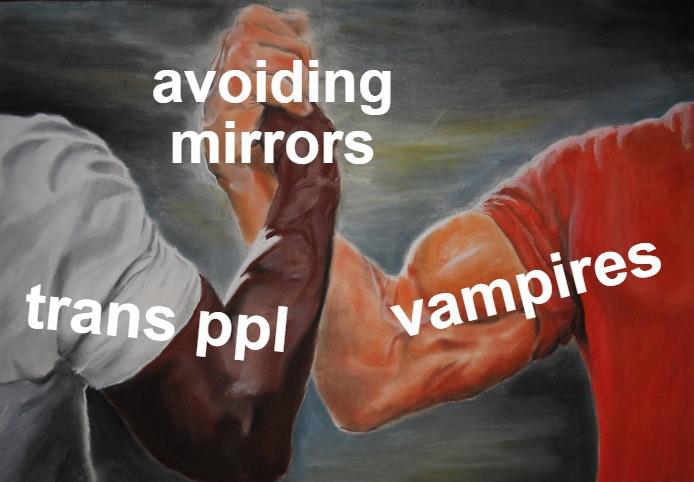}
  		\caption{\scriptsize {Real:\{1, 2, 2, 1\} \\ Predction:\{0, 2, 0, 1\}}}\label{fig:53a}		

	\end{subfigure}
	\hspace{2em}
        \begin{subfigure}{0.16\textwidth}
            \captionsetup{justification=centering}
		\includegraphics[width=32mm, height=32mm]{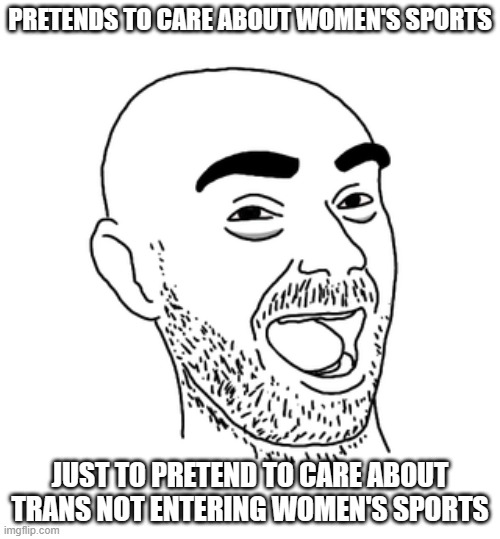}
  		\caption{\scriptsize {Real:\{1, 1, 2, 1\} \\  Prediction:\{1, 2, 2, 1\}}}\label{fig:53b}
              
	\end{subfigure}
\caption{Examples of memes misclassified by MemeCLIP across four tasks. The labels are in the form of \{Hate, Target, Stance, Humor\}. Label details are outlined in Figure \ref{fig:examples_1}.}

 \label{fig:misclassified}
\end{figure}

We present two examples of images misclassified by MemeCLIP in Figure \ref{fig:misclassified}. The meme presented in Figure \ref{fig:53a} is hateful and opposes the values of the LGBTQ+ Pride movement under the guise of benign imagery and text, but is misclassified as non-hateful and neutral. Figure \ref{fig:53b} shows a hateful meme mocking an individual, but MemeCLIP classifies it as hate against a community since the text mentions communal words such as "trans" and "women". Both memes were correctly classified as humorous by the model.

\section{Conclusion}

In this work, we release PrideMM, a multimodal and multi-aspect dataset comprising 5,063 memes related to the LGBTQ+ movement, addressing a serious gap in data resources in this domain. This dataset provides memes annotated across three aspects and one sub-class, allowing for greater flexibility in the establishment of ethical guidelines by social media policymakers. We further introduce MemeCLIP, a lightweight yet effective framework to harness CLIP's knowledge for multimodal meme classification while not being reliant on extraneous models to create augmented data for training. 

With each module of MemeCLIP, we tried to address pervasive problems in the domain of hateful image classification. Specifically, we used a cosine classifier to counter class imbalances, which are common in multi-aspect datasets with many classes for each aspect. We utilize feature adapters to mitigate overfitting as datasets in this domain have a relatively small scale. We further use linear projection layers to dissociate the image and text modalities in the representations created by the pre-trained CLIP encoders as the pre-training dataset of CLIP majorly consists of image-text pairs that convey the same overall meaning, which may not be the case with creative or sarcastic memes. We show that CLIP, one of the most basic multimodal models, when combined with lightweight additional modules, can compete with or outperform models that require extraneous models to create augmented data, or even VLMs such as GPT-4. Our work is grounded on the importance of a data-centric approach to solving problems in this domain, rather than creating larger frameworks and incorporating extraneous data. 

Our dataset endeavors to foster a deeper understanding of online interaction, community-building, and social change, whereas our framework is a step toward effective content moderation that helps create an inclusive, diverse, and equitable digital environment for all.

\clearpage

\section*{Ethical Considerations}

\textbf{User Privacy.} \quad Our dataset only comprises text-embedded images collected from publicly accessible web pages with no inclusion of user data. To the best of our knowledge, there are no copyright concerns associated with them. Since our OCR tool may have potentially scraped user identifiers such as Twitter usernames containing `@', we removed such data from the text during the pre-processing step. 

\noindent \textbf{Biases.} \quad Due to the size of the dataset and the number of annotation tasks, we acknowledge that some data samples may be misclassified, and not all annotators would agree on the same labels for a given sample. Further, due to the subjective nature of this topic, unintentional biases in the dataset's distribution and annotation may exist. We endeavored to minimize such occurrences by using comprehensive annotation guidelines and a three-phase annotation schema.

\noindent \textbf{Potential Risks.} \quad PrideMM contains images conveying targeted hate towards specific individuals, communities, ethnic groups, and other entities. While we release our dataset to help robustify content moderation and foster safer spaces online, this data may potentially be used to spread hate and discrimination. We ask researchers to be aware that the inherent biases present in the dataset may negatively influence hate speech detection and moderation methods. Further, we release our multi-aspect dataset in order to give social media policymakers more freedom to censor specific types of harmful memes. However, we acknowledge that this may also lead to over-moderation and negatively affect social media users' freedom of expression.

\noindent \textbf{Annotation.} \quad Five annotators were hired to annotate the images for our dataset. The annotators were Indian university students aged 20-25 and were familiar with the LGBTQ+ Pride movement. They were compensated fairly as per the standard local rate. Given the nature of this study, we acknowledged that the annotators may find the images disturbing or distressing. The annotators were given the option to opt out of the annotation process at any given time.

\noindent \textbf{Reproducibility Statement.} \quad We provide implementation details and hyperparameter configurations for all the implemented models in Appendix \ref{appendix:implement}. The PrideMM dataset, source code for MemeCLIP, and MemeCLIP's pre-trained weights are publicly available at \href{https://github.com/SiddhantBikram/MemeCLIP}{https://github.com/SiddhantBikram/MemeCLIP}.

\noindent \textbf{Environmental Impact.} \quad Leveraging hardware such as GPUs to train deep learning models is known to have a significant environmental footprint, primarily due to their high energy consumption resulting in carbon emissions. To mitigate the environmental impact of our research, we adopted a fine-tuning technique for pre-trained deep-learning models. This method allowed our models to generalize faster on new datasets, resulting in fewer computational resources consumed. Once these models are trained using GPUs, they can be loaded on relatively lightweight CPUs for inference purposes, attenuating potential environmental consequences.

\section*{Limitations}

The proposed dataset PrideMM encompasses memes posted from 2020 to 2024, representing a snapshot of social media interactions within this specific period, which may fail to capture the dynamics of the LGBTQ+ movement on social media over an extended timeframe. Further, due to the subjective nature of the LGBTQ+ Movement, the size of the dataset, and the number of annotators, the annotation process is inherently prone to biases. While we include images from multiple sources in our dataset, we acknowledge the limited scope of our dataset compared to the vast number of social media platforms. Further, the intricate nature of memes may not be completely captured by our aspects, and more domain-specific aspects may be used to capture the context better. Due to the limited size and availability of labeled datasets in this domain, supervised frameworks such as MemeCLIP may be surpassed by unsupervised methods such as LLMs for content moderation as the capabilities and model size of LLMs progress. Finally, MemeCLIP and other models trained on PrideMM and similar datasets may exhibit biased predictions due to the presence of unintentionally biased data.

\bibliography{custom}

\appendix

\section{Appendix}

\subsection{Topic Modeling} \label{appendix:topicmodeling}

We applied the Sparse Additive Generative Models of Text (SAGE) \cite{eisenstein2011sparse} topic modeling technique to identify noteworthy words across various class labels within our dataset. We set the hyperparameters max\_vocab\_size to 1000 and base\_rate\_smoothing to 1. Tables \ref{table:topic1} and \ref{table:topic2} present the most notable words for each class as identified by SAGE along with their corresponding salience scores. Among hate targets, words like `rowling', `shapiro', `conservative', `gays', `bethesda', and `corporations' are assigned high scores by SAGE, helping identify the most targeted entities for each label. Within samples labeled \textit{Support}, words such as `comfortable', `expression', and `supportive' hold relevance as they convey acceptance and support.

\begin{table*}[!h]
\centering
\scriptsize
\caption{Topic Modeling for Hate and Target classification tasks. We report the top 5 words for every label in each task sorted according to their salience score.}
\begin{tabular}{c c | c c c c}
\hline \hline
\multicolumn{2}{c|}{\textbf{Hate}} & \multicolumn{4}{c}{\textbf{Target}} \\
\textbf{No Hate} & \textbf{Hate} & \textbf{Undirected} & \textbf{Individual} & \textbf{Community} & \textbf{Organization}\\
\hline \hline

brooke (0.919) & woke (0.306) & father (1.111) & rowling (2.122) & transphobes (1.002) & bethesda (2.649) \\

envy (0.917) & conservative (0.286) &  oppression (1.053) & shapiro (1.814) & terfs (0.934) & corporations (2.334) \\

comfortable (0.859) & children (0.279) & event (1.043) & ben (1.754) & conservatives (0.846) & companies (2.322) \\

expression (0.842) & warning (0.279) & center (0.932) &biden (1.733) & turning (0.759) & disney (2.265) \\

subscribers (0.820) & marriage (0.204) & bigot (0.920) & walsh (1.595) & gays (0.758) & russia (2.232) \\

\hline \hline
\end{tabular}

\label{table:topic1}
\end{table*}

\begin{table*}[!htpb]
\centering
\scriptsize
\caption{Topic Modeling for Stance and Humor classification tasks. We report the top 5 words for every label in each task sorted according to their salience score.}
\begin{tabular}{c c  c | c c}
\hline \hline
\multicolumn{3}{c|}{\textbf{Stance}} & \multicolumn{2}{c}{\textbf{Humor}} \\
\textbf{Neutral} & \textbf{Support} & \textbf{Oppose} & \textbf{Humor} & \textbf{No Humor}\\
\hline \hline

envy (1.012) & comfortable (1.073) & warning (0.629) & envy (0.480) & risk (0.826) \\

min (0.964) & subscribers (1.032) & walsh (0.624) & femboy (0.418) & comfortable (0.723) \\

content (0.938) & brooke (1.000) & matt (0.615) & miss (0.352) & walsh (0.720) \\

thinks (0.845) & expression (0.946) & replies (0.601) & mematic (0.342) & youth (0.720) \\

republican (0.804) & supportive (0.907) & oppressed (0.599) & thinks (0.333) & protect (0.640) \\

\hline \hline
\end{tabular}

\label{table:topic2}
\end{table*}

\subsection{Data Filtering}
\label{apx:filter}

 We screened the images according to the following criteria:

\begin{itemize}
    \item \textbf{Irrelevant Images}: We curated images relevant to the LGBTQ+ movement and discarded non-relevant images.
    \item \textbf{All Text or no Text Images}: We discarded images that did not contain significant visual content or did not have any embedded text. 
    \item \textbf{Non-English Text}: We majorly collected images that had English content, however, some images may contain non-English words. Our OCR tool was set to English ensuring that only English text was extracted from images. 
    \item \textbf{Low-Quality Images}: We discarded images that were highly distorted, blurred, or degraded. We also removed images with illegible text.
\end{itemize}

\subsection{Implementation Details} \label{appendix:implement}

We conducted all our experiments on Pytorch 2.1.2 combined with an NVIDIA Tesla T4 GPU with 16 GB of dedicated memory. We set the batch size to 16 and trained each model for 10 epochs while monitoring validation AUROC to save the best model for each run. Under these settings, training and validating MemeCLIP for one epoch takes 12 minutes and occupies 7 GB of dedicated memory. 

We empirically found the most optimal learning rate for each model. We used a learning rate of $10^{-5}$ for ViT-L/14 and CLIP Image-Only. We used a learning rate of $5 \times 10^{-5}$ for BERT and CLIP Text-Only. We used a learning rate of $10^{-3}$ for CLIP and CLIP-Adapter. For MOMENTA, HateCLIPper, and ISSUES, we used the default settings set by their respective authors. For MemeCLIP, we set the learning rate to $10^{-4}$. We set the scaling factor $\sigma$ for the cosine classifier to 30, and the residual ratio $\alpha$ for the Feature Adapters to 0.2. 

Our framework, MemeCLIP, was built upon the base CLIP model provided by OpenAI's official CLIP library\footnote{https://github.com/openai/CLIP}. We also implemented CLIP, CLIP Image-Only, and CLIP Text-Only models by using this library. We implemented the Visual Transformer model by using the timm library provided by Huggingface\footnote{https://github.com/huggingface/pytorch-image-models}. We implemented BERT by using the Huggingface Transformers library\footnote{https://github.com/huggingface/transformers}. We obtained the code released by the authors of CLIP-Adapter\footnote{https://github.com/gaopengcuhk/CLIP-Adapter}, MOMENTA\footnote{https://github.com/LCS2-IIITD/MOMENTA}, HateCLIPper\footnote{https://github.com/gokulkarthik/hateclipper}, and ISSUES\footnote{https://github.com/miccunifi/ISSUES} to test their respective methods. The total number of parameters for each method is listed in Table \ref{table:parameters}.

\begin{table}[!h]
\setlength{\tabcolsep}{3pt}
\renewcommand{\arraystretch}{1.2}
\centering
\caption{Number of Parameters for each implemented method.}

\begin{tabular}{ c  c }
\hline \hline
\textbf{Method} & \textbf{Number of Parameters (M)}\\

\hline \hline
ViT-L/14 & 307 \\
CLIP Image-Only & 307\\
CLIP Text-Only & 123\\
\hline
CLIP ViT-L/14 & 428\\
CLIP-Adapter & 428\\
MOMENTA & 358\\
HateCLIPper & 430 \\
ISSUES & 452\\
\hline
MemeCLIP & 430\\
\hline \hline
\end{tabular}
\label{table:parameters}
\end{table}

\end{document}